# Hold the Suspect!
## An Analysis on Media Framing of Itaewon Halloween Crowd Crush

**TaeYoung Kang**
Underscore, KAIST
minvv23@underscore.kr

**Abstract**

Based on the 10.9K articles from top 40 news providers of South Korea, this paper analyzed the media framing of *Itaewon Halloween Crowd Crush* during the first 72 hours after the incident. By adopting word-vector embedding and clustering, we figured out that conservative media focused on political parties' responses and the suspect's identity while the liberal media covered the responsibility of the government and possible unequal spillover effect on the low-income industry workers. Although the social tragedy was not directly connected to institutional politics, the media clearly exhibited political bias in the coverage process.

## 1 Introduction

On October 29, 2022, a huge accident occurred in the narrow alleys of Itaewon, Seoul, South Korea. As many people have been gathering to enjoy the Halloween festival, 159 people injured and 196 people killed due to the crowd crush. This was the largest incident of crowd-caused accidents in Korea in the past 70 years, and due to the temporal and spatial specificity of Itaewon and Halloween, there were also many victims who were not locals; 16.4% of the fatalities were foreigners.

Due to the severity and the unprecedented nature of *Itaewon Halloween Crowd Crush*, the majority of Korean newspaper coverage in the days following the accident was related to this tragedy. In the case of global news outlets, they were relatively free from local political disputes in Korea and were able to conduct in-depth reporting from a third-party perspective. However, in the case of local media, division occurred during the reporting process depending on the political inclination of the news providers. In this study, we will analyze the *media framing* that appeared differently according to the political stance in relation to the crowd crush of Itaewon.

## 2 Research Objectives

In the field of political communication, along with agenda-setting and priming, *framing* is a term depicting a strategy to politically emphasize certain aspects and minimize or ignore others, so that the media can affect how the public perceives the causes, effects, and responses to the event (Scheufele, 2000; Groeling, 2013; Dunway and Graber, 2022). Although we cannot completely rule out the possibility that the level of emphasis on the identical issue might differ by tangible political status (Barkemeyer et al., 2017), even when the social catastrophes and natural disasters are not directly related to political figures and institutions, the public can still perceive the identical events in a different way (Houston et al., 2012; Yihong and Boin, 2020; Triantafillidou et al., 2022).

The political framing of media gets even more 'effective' as people's preexisting beliefs can also influence how they interpret and react to different frames presented by the partisan news outlets; this phenomenon is commonly referred to in several adjacent terms including *polarization, selective exposure of news media, echo chamber*, and *filter bubble* (Stroud, 2008; Knobloch-Westerwick, 2014; Duboi and Blank, 2018; Bakshy et al., 2015; van der Meer et al., 2022). Considering this feedback loop of news consumption, researchers should investigate the landscape of media frames describing social catastrophes and disasters as they can shape the *public memory*, providing the basis for the attribution, understanding of the cause and effect, evaluation of the person or institution in charge of a given incident.

The tragedy of Itaewon would not be an anomaly. The incumbent president from the conservative party was suffering from low approval ratings of around 32% during the last week of October, and thus, it could provide an opportunity for partisan media to exploit the accident as a means of political confrontation. Liberal media outlets may focus on the human impact of the disaster, such as the shortcomings in government planning and management, emphasizing the need for stronger regulations and oversight to prevent future crowd crush. Conversely, conservative media may depict this event as a natural occurrence that could not have been easily prevented through governmental preparedness, and therefore, downplay any broader systemic issues or lack of regulations.

Despite the fact that this general argumentative pattern could have persisted for a long time, the alignment of the frame was established relatively early on. On Tuesday, November 1st, it was revealed that warnings of potential crowd danger were continuously reported to the police, hours before the accident, and thus after 72 hours from the initial tragedy, front page of major newspapers were all dominated by this disclosure. If we think in reverse, in the absence of objective evidence, the first few days of coverage by different media outlets may have a higher likelihood of containing a clearer political bias. By analyzing these articles, we can understand how different political speakers frame seemingly apolitical social events in the absence of specific evidence. Through this, we not only aim to grasp the discourse of the Itaewon Halloween Crowd Crush in the local context, but also to make additional discoveries that have not been captured in conventional analysis of media coverage bias.

## 3 Data and Methods

### 3.1 Data

We collected every news article uploaded in the news portals of South Korea between October 29th and October 31st and filtered them based on following conditions. First, the title or content of a news should include either one of the following words; Itaewon 이태원, Halloween 할로윈, disaster 참사, crowd crush 압사, and mourning 애도. Second, only the articles from news outlets that are included in the top 40 news release frequency during the data collection period should be considered for the representativeness. Based on these criteria, we used 10,908 news articles in the media framing analysis. However, for the purpose of more stable development of the language model, we also collected an additional 13,762 popular news articles written in 2022 just for the use in model training.

### 3.2 Methods

To understand the topics of news articles about the Itaewon disaster, we clustered them through following procedures.

First, based on 24.7K news articles consisted of 10.9K crowd crush coverage and additional 13.8K news text, we trained FastText word embedding. We decided to use a lightweight model, which may not be the latest but is more suitable for our task of sociological interpretation instead of the larger pretrained language model, as they tended to suffer from capturing the delicate expressions on the social disaster coverage.

Second, the average embedding for each news document was calculated and the K-means algorithm was applied to cluster the articles into 15 different categories. The optimal number of clusters was determined using the Silhouette Score, which was tested in the range between three and eighty.

Third, to obtain only the essential news articles for each cluster, half of the text closest to the centroid of each cluster was analyzed. This approach minimized the potential noises in the dataset and enhanced the interpretability for human researchers at the same time.

## 4 Analysis

### 4.1 Summary Statistics

The topics of the 15 clusters are as follows, and the main keywords extracted by the TextRank algorithm for each cluster can be found in the Appendix.

1) Missing person registration
2) Lack of preventive measures
3) Joint memorial service arrangements
4) Memorial service arrangements at Itaewon
5) Central Disaster Countermeasures Headquarter (CDCH) meeting
6) Cancellation of cultural events during the national period of mourning.
7) Number of casualties
8) Response by local government organizations
9) Breaking news on the day of the accident
10) On-site investigation by the National Forensic

Service(NFS)
11) Memorial statement by the head of the local government organization
12) Description of the accident scene
13) Response by the President
14) Investigation of the "suspects"
15) Response by political parties.

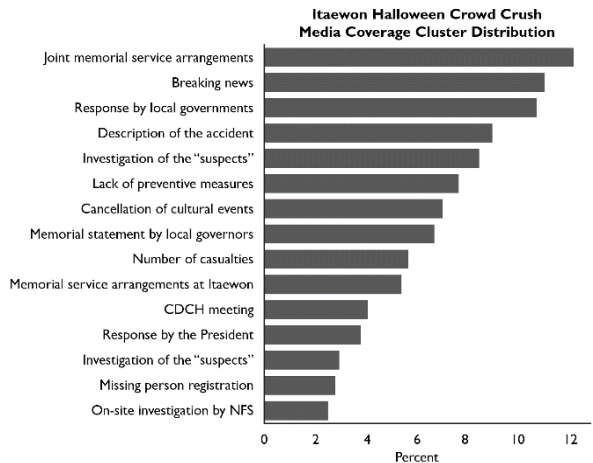

Figure 1 : Percentage of News Clusters

The most reported topic was *(3) Joint memorial service arrangements*, accounting for 12.2% of all news. In the case of articles of *cluster (2)* criticizing the lack of prior measures, there was a middle level occupancy rate of 7.7%, and articles related to the investigation of causes and responsibilities, CCTV footage analysis, "a suspect deliberately pushing the crowds" that belonged to *cluster (14)* accounted for a very low proportion of 3.0% in total data.

To facilitate an understanding of the progression of Itaewon issue in South Korea, illustration on specific topics will be needed. In regard to *(6) Cancellation of cultural events during the national period of mourning*, the president declared Itaewon as a special disaster area and designated a week as the national period of mourning. As a result, cultural events hosted by public institutions were significantly reduced, and private events were also advised to be canceled. *(14) Investigation of the "suspects"* pertains to a group of individuals who were commonly captured in CCTV footage and videos uploaded onto social media on the day of the incident. Initially, it was suggested that this group had initiated the pushing of victims, thus leading to the occurrence of the accident. However, it was later determined that this was simply a rumor, and they were cleared of any charges through the investigation process.

### 4.2 Main Analysis

By filtering 22 major media outlets that are considered to have clear political leanings once again, we calculated the average political slant by each topic clusters. Liberal, moderate, and conservative news providers were weighted as -1, 0, and 1 respectively, and labeled based on the majority voting from 10 journalists. The list of media outlets by political slant is stated in Appendix. Since the baseline of average political distribution is (-1×6+0×9+1×7)/3=0.333, the political slant per topic visualized in horizontal axis of Figure 2 was normalized by this value.

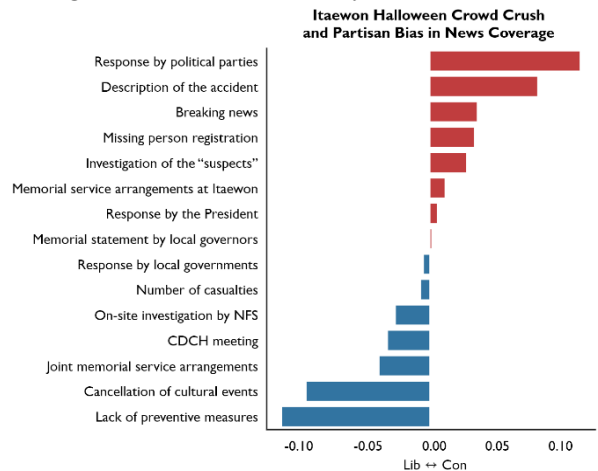

Figure 2 : Partisan Bias in News Coverage

The results of the analysis indicate a distinct differentiation of topics based on the political leanings of the media. As can be observed in the bar graph below, the longer the bar extends in a positive direction, the greater the coverage ratio of conservative media on the topic. Conversely, the longer the bar extends in a negative direction, the greater the coverage ratio of liberal media on the topic.

Conservative-leaning media outlets focused on the topics of *Response by political parties* and *Description of the accident scene*, while liberal media outlets primarily covered *Lack of preventive measures* and the sudden *Cancellation of cultural events during the national period of mourning*. Additionally, the topic of *Investigation of the "suspects"* regarding the spread of rumors about the individual responsible for the tragedy and the investigation conducted by the prosecution and police on the group that began to push the crowd, was also covered more extensively by conservative media outlets. Interestingly, when excluding the topic of *Lack of preventive measures*, other articles tended to focus on delivering dry information rather than making strong criticisms. For example, the topic of *Response by political parties* was covered extensively by conservative media, however, the number of articles that treated the Democratic Party negatively was not significant. Similarly, the liberal media covered the second most *Cancellation of cultural events during the national period of mourning*, but the proportion of articles that directly addressed the livelihood issues of low-income cultural industry workers was not high. This suggests that the media may implement framing strategies by selectively adjusting the proportion of information that is exposed to readers rather than clearly expressing their own partisan opinions, based on their own political tendencies.

## 5 Conclusion

The *Itaewon Halloween Crowd Crush* of South Korea was not only an unprecedented tragedy, but also provided an opportunity to investigate how a seemingly apolitical disaster can be reported differently depending on the political leanings of media outlets. In other words, our study is both a traditional social science research on media framing and an NLP-applications on large-scale text data. We collected 10.9K news articles on the accident written within the first 72 hours after the outbreak. By combining language embedding and clustering, we were able to observe a clear topic differentiation depending on the media's political leanings.

Some might expect that liberal media would have a greater likelihood of politicizing the incident, as the crowd crush occurred during a time when a conservative incumbent party was suffering from low approval ratings. However, it was actually conservative media that most heavily covered the topic of *Response by political parties*. This implies that politically ambivalent attitude can be a media strategy to dilute the administrative responsibility, instead of standing up for the incumbents in a obvious way. Furthermore, it was found that the coverage of suspects, who were ultimately cleared of guilt later, was also covered more actively by the conservative media. It can be understood within the context of the traditional conservative rhetoric that tends to *privatize* certain events. On the other hand, liberal media that were critical of the incumbent administration focused on *Lack of preventive measures* and concentrated on reporting the response of official institutions such as National Forensic Service and Central Disaster Countermeasures Headquarter. Additionally, the fact that the coverage of the impact on livelihoods of cultural industry workers, who were affected by the national period of mourning is related with the progressive tradition that highlights the unequal effects of identical social incident. In summary, although this catastrophe has no direct connection to institutional politics, the media clearly exhibited political bias in the coverage process.

## Limitations

This study analyzes the different framing strategies of the media according to their political inclinations, however, it does not go as far as to examine how the same concept is used differently by distinct news outlets. The convergence of media position of individual news providers that varies over time is also not discussed in the paper. This is a limitation of the research methodology, but at the same time, it is also an inevitable result due to the rapid convergence of local opinion within the first 72 hours. Therefore, future research could attempt a more long-term examination of media strategy changes on similar cases, and a more sophisticated engineering approach utilizing advanced language models.

## Ethics Statement

The annotators who participated in the political leaning labeling of the media outlets, independently provided their own opinions on the given newspaper companies without prior knowledge of the research purpose. Additionally, as this study deals with article-level analysis, it does not compromise any personal information.

## Acknowledgments

I would like to extend my thanks to Media Today (http://www.mediatoday.co.kr/) for their support, and to journalist Geum Joon-kyung (금준경) for his valuable advice that enriched my research idea. Their contributions greatly aided the development of this paper.

## A  Appendices

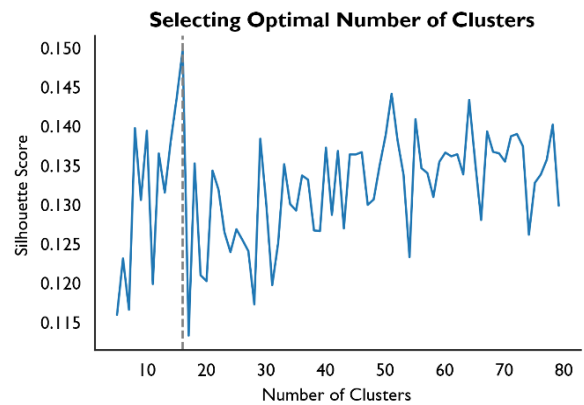

Figure A1 : Selecting Optimal Number of Clusters

| No. | TextRank Keywords |
|---|---|
| 1 | missing person, report, center, Seoul, Yongsan-gu, Seoul City, missing, related |
| 2 | police, accident, safety, management, deployment, minister, personnel, measure |
| 3 | funeral home, joint, preparation, death, accident, Seoul, journalist, funeral, citizen |
| 4 | Seoul, Yongsan-gu, memorial, accident, scene, victim, Itaewon Station, occurrence |
| 5 | accident, president, meeting, government, minister, presence, Prime Minister |
| 6 | event, festival, cancellation, mourning, scheduled, safety, national, postponement |
| 7 | death, hospital, confirmation, funeral, identity, Seoul, mortuary, accident, transfer |
| 8 | relief, occurrence, emergency, Seoul, death, damage, disaster, meeting, government |
| 9 | scene, area, crowd, scale, fire department, police, Halloween, death, damage |
| 10 | scene, investigation, Investigation Headquarters, Seoul Metropolitan Police |
| 11 | mourning, funeral home, victim, joint, related, relief, death, memorial, scene |
| 12 | person, citizen, crowd, police, video, time, friend, professor |
| 13 | president, relief, mourning, Yoon Seok-yeol, national affairs, condolence |
| 14 | investigation, identity, composition, analysis, examination, cause, CCTV |
| 15 | People's Power Party, Democratic Party, representative, National Assembly, member |

Table A1: Cluster Keywords

| Slant | List of News Providers |
|---|---|
| Lib | JTBC, MBC, 경향신문, 오마이뉴스, 프레시안, 한겨레 |
| Moderate | KBS, SBS, YTN, 뉴스 1, 뉴시스, 서울신문, 연합뉴스, 연합뉴스 TV, 한국일보 |
| Cons | 데일리안, 동아일보, 매일경제, 조선일보, 중앙일보, 채널 A, 한국경제 |

Table A2: Labeling the News Outlets